\documentclass[11pt]{article}
\pdfoutput=1
\usepackage[]{EACL2023}

\usepackage{times}
\usepackage{latexsym}
\usepackage{amsmath}
\DeclareMathOperator*{\argmax}{arg\,max}

\usepackage{todonotes}

\usepackage[T1]{fontenc}
\usepackage[utf8]{inputenc}

\usepackage{microtype}

\usepackage{inconsolata}
\usepackage{color}

\title{Noisy Channel for Automatic Text Simplification}

\author{\ Oscar M. Cumbicus-Pineda$^{1}$, \ Iker Gutiérrez-Fandiño$^{2}$, \ Itziar Gonzalez-Dios$^{3}$, \ Aitor Soroa$^{3}$ \\
  $^1$Ixa group, UPV/EHU, UNL\\
  $^2$University of Deusto\\
  $^3$HiTZ Center - Ixa, University of the Basque Country UPV/EHU \\
\texttt{ocumbicus001@ikasle.ehu.es}\\
\texttt{ikergutierrez@opendeusto.es}\\
\texttt{\{itziar.gonzalezd,a.soroa\}@ehu.eus}
}

\begin{document}
\maketitle
\begin{abstract}

  In this paper we present a simple re-ranking method for Automatic
  Sentence Simplification based on the noisy channel scheme. Instead
  of directly computing the best simplification given a complex text,
  the re-ranking method also considers the probability of the simple
  sentence to produce the complex counterpart, as well as the
  probability of the simple text itself, according to a language
  model. Our experiments show that combining these scores outperform
  the original system in three different English datasets, yielding the best
  known result in one of them. Adopting the noisy channel
  scheme opens new ways to infuse additional information into ATS
  systems, and thus to control important aspects of them, a known
  limitation of end-to-end neural seq2seq generative models.
\end{abstract}

\section{Introduction}
\label{sec:introduction}

Automatic Text Simplification (ATS) aims to rewrite text into a form
that is easier to understand, while retaining the original
meaning. Its is an active area of research with many interests in improving web accessibility \cite{Alarcon2021}, for cognitive disabled users \cite{kamran2022web,moreno2021designing}, for scientific information access \cite{Ermakova}, and in general it is intended for social good \cite{SocialGood}.

In the last years there has been a remarkable advance in ATS due to
the advent of deep learning techniques and neural language models. This has lead to a variety of methods that
follow the seq2seq architecture to build models that estimate $p(y|x)$, the probability of producing a
simple sentence $y$ given the complex sentence $x$ \cite{nisioi2017exploring,zhang-lapata-2017-sentence,martin2020multilingual,Lin_Wan_2021,Omelianchuk}. One of the main problems of this approach is the lack of control mechanisms to prevent the system to hallucinate or produce repetition when generating $y$, as well to infuse new information into the system. Previous works have tried to control the generation of simple sentences by using control codes that condition the system output with variables such as length, number paraphrases, lexical complexity, and syntactic complexity, etc \cite{martin-etal-2020-controllable,sheang-saggion-2021-controllable}. Other works such as \citet{Clive-2021} use control prefixes, including input-dependent conditional information pre-trained models, incorporating learnable attribute-level representations at different layers of a transformer.

In this paper we propose an alternative method to control the generated simplifications of a seq2seq ATS model. Instead of controlling the simplifications at decoding time, we propose to re-rank the top candidates of an neural seq2seq model according to the noisy channel scheme  ~\cite{10.5555/972470.972474,yee-etal-2019-simple,Yu-Lei}. The noisy channel scheme decomposes $p(y|x)$ into $p(x|y)$, the probability of generating the complex sentence $x$ given the simple sentence $y$, and $p(y)$, the probability of the simple sentence $y$. We present a re-ranking method that considers all three probabilities using different models. Because each of the models is independently trained on different data, our method is able to infuse additional knowledge into a traditional seq2seq ATS system. The noisy channel method has been mainly applied in machine translation, but it was also applied for Document Compression \cite{daume2002noisy}, exactly in the dropping word task (related to summarization). Dropping or deleting words is also one of the operations an ATS systems should carry our, but it is not limited to that: ATS systems should also split sentences, reorder words or sentences and add necessary words or phrases. Moreover, ATS systems's output should be coherent, and by  only dropping words coherence can become deficient.

The contributions of this paper are the following:
\begin{itemize}
\item We propose a re-ranking method for seq2seq ATS systems based on noisy channel.
\item We show that our method outperforms the original system in three
  different English datasets, yielding the best known result in one of them.
\item We show that the method is able to infuse new information in neural
  ATS systems, as the additional models in the noisy channel scheme can be
  trained on complementary data.

\end{itemize}

\section{Noisy Channel based re-ranking for ATS}
\label{sec:noisy-channel-ats}

Given a complex text $x = \{x_1,\ldots,x_M\}$ and its simplified version
$y = \{y_1,\ldots,y_N\}$, a seq2seq ATS system builds a model that estimates
$p(y|x)$. At inference time, and given the input $x$, the model is used to
select the simplification that maximizes this probability, e.g.,
$\argmax_y p(y|x)$. The noisy channel scheme is an alternative that has been widely
used in machine translation, and which uses Baye's rule to parametrize
$p(y|x)$ as $\frac{p(x|y)p(y)}{p(x)}$. Because we are interested in finding
$y$ that maximizes the probability, the denominator $p(x)$ is
discarded.
% , as shown in equation \ref{eq:bayes}.
% \begin{equation}
%     \label{eq:bayes}
%     argmax_y \frac{p(x|y)p(y)}{p(x)} = argmax_y
%   p(x|y)p(y)
% \end{equation}

In this paper we explore using the noisy channel scheme to re-rank the
candidates generated by the seq2seq ATS model. Specifically, we
propose to build three separate models to estimate $p(y|x)$, $p(x|y)$
and $p(y)$. Following usual practice, will be henceforth refer to
these models as the \emph{direct model}, \emph{channel model} and
\emph{language model}, respectively. Given $x$, our system first
retrieve the list of top-k candidates given by the direct model, as
well as the corresponding $p(y|x)$ probabilities. Then, the channel
and language models are used to compute $p(x|y)$ and $p(y)$ for each
$(x,y)$ pair. Finally, the top-k candidates are re-ranked according to
a combination of the three probabilities, using the same score
function as in \cite{Yu-Lei}\footnote{We tried alternative score
  functions such as the proposed in \cite{yee-etal-2019-simple} with
  similar results.}:
\begin{equation}
  \label{eq:eq1}
  \begin{split}
    &\lambda_1 \log p(y|x) + \lambda_2 \log p(x|y)\\
    &+ \lambda_3 \log p(y) + \lambda_4\cdot N
  \end{split}
\end{equation}

where $N$ is the length of sentence $y$ and
$\lambda_1,\lambda_2,\lambda_3, \lambda_4$ are hyper-parameters. As in \cite{Yu-Lei}, we use $\lambda_4$ to penalize the model for
generating simple sentences that are too short.

\section{Experimental setup}
\label{sec:experimental-setup}

This section describes the experiments we conducted to evaluate the effectiveness of the proposed noisy channel re-ranking method for English ATS. We describe the experimental setup, including the built models, datasets and evaluation metrics, as well as a description of the different systems and baselines.

%The experiments we performed were for monolingual simplification in English, for this we trained our models using a transformers seq2seq architecture, on an NVIDIA A100 GPU, the approximate training time for each model was 24 hours, here are the details.

\subsection{Building the models}
\label{sec:models}

\textbf{The direct model} $p(y|x)$ is based on BART large, a
encoder-decoder language model with 400M parameters that is
pre-trained on English text~\cite{lewis2020bart}. The model is
fine-tuned using the training splits of the respective datasets. We
used default values for the learning rate ($5\times10^{-5}$ with a
linear schedule and a warmup of $500$ steps). We trained the model for
10 epochs, and select the checkpoint with smallest loss on the
respective validation sets. To generate the top-$k$ candidates, we used beam search with a beam size of $k$, and set the maximum sentence length of $100$. 

%To generate the simplifications we set the decoder of our model with the number of beams equal to 10, maximum length to 100, the parameter num\_return\_sequences which is the number of highest scoring beams to be returned was set to 10, the early stopping, the return dict in generate and the output scores to True.

\textbf{The channel model} $p(x|y)$ is trained using the exact same
settings as the direct model, but when fine-tuning the simple and
complex sentences are swapped, so that the model learns to produce
complex sentences given the simpler ones.

\textbf{The language model} $p(y)$ is based on Bert-large-uncased \cite{Bert-large-uncase}. We
continue to train the pre-trained model using the Simple Wikipedia,
and the default language modeling objective. Again, we used default values: a weight decay of $1\times10^{-2} $, learning rate of $ 2\times10^{-5}$ with warmup of $500$ steps, and vocabulary size of $30,522$. Not being a generative model, BERT does not directly compute
$p(y)$ for any given $y$. Thus, we estimate this probability at inference time by
consecutively masking each word in $y$ and calculating the probability
that BERT produced the correct word in the corresponding
position. Specifically, we estimate $\log p(y)$ using the following
equation:
\begin{equation}
  \label{eq:bert}
  \log p(y) \sim \sum_i^N \log p_{\mathrm{bert}}(y_i|y^{(i)})
\end{equation}

where $y_i$ is the $i$th token, $y^{(i)}$ is the original sentence
with the $i$th token replaced by the \texttt{MASK} token, and $N$ is
the length of the sentence.

\subsection{Datasets}
\label{sec:datasets}

We perform our experiments using three English ATS
datasets: WikiSmall, WikiLarge/TurkCorpus, and Newsela. WikiLarge and WikiSmall were built by automatically aligning sentences
belonging to the same article in English Wikipedia and Simple English
Wikipedia. In both cases, we used the splits provided by
\citet{zhang-lapata-2017-sentence}. The WikiSmall dataset contains
$88,837$ training instances, $205$ validation instances, and $100$
test instances. Regarding WikiLarge, the training set contains
$296,402$ instances and the validation set $2,000$. For evaluation, we
used the TurkCorpus dataset \cite{xu-etal-2016-optimizing}, which
comprises eight manually generated simplifications for each instance
in the WikiLarge test set. Finally, the Newsela dataset consists more
than $1k$ news articles that were rewritten four times at different
complexity levels. We used the train/development/test splits from
\cite{xu-etal-2015-problems}, containing $94,208$/$1,129$/$1,076$
sentences respectively.

\subsection{Hyperparameter search}
\label{sec:hyperp-search}

We perform a hyperparameter grid search on the respective development
set to find the optimal values for
$\lambda_1, \lambda_2, \lambda_3, \lambda_4$ in Eq. \eqref{eq:eq1}. We
tried all combinations ranging the value of each $\lambda$ between $0$
and $1$, with $0.1$ increments. The combination that obtains the best
SARI score (see next section) in the development set is selected, and
used in the respective test sets. We used $k=10$ in our experiments.

\subsection{Evaluation Metrics and Systems}
\label{sec:exper-sett}

Following usual practice, we report the results using SARI~\cite{xu-etal-2016-optimizing} and Flesch–Kincaid Grade Level (FKLG) readability metric \cite{kincaid1975derivation}. We make the experiments with the original \textbf{BART} system, as well as
our \textbf{NC-TS} system, the result of selecting the candidates
after re-ranking based on the noisy channel scheme. We also test  an \textbf{Oracle} system in the development test, which is
the result of selecting the candidate that maximizes the SARI score
wrt. the reference sentence in the gold standard, and represents an
upperbound of the re-ranking process. Finally, we include a baseline (dubbed \textbf{Cosine}) that re-ranks the candidates according to their cosine similarity wrt. the complex sentence. Sentence embeddings are calculated using Sentence Transformers \cite{reimers-2019-sentence-bert}\footnote{We used the pre-trained \textrm{paraphrase-albert-small-v2} model.}.

%system based on the Sentence Transformers \cite{reimers-2019-sentence-bert} framework to perform a simple ranking, sending for each complex-simple sentence pair and calculating the similarity between the two with the cosine similarity loss function and the pre-integrated paraphrase-albert-small-v2 model. 

\section{Results}
\label{sec:results}

\begin{table}[t]
  \centering
  \small
  % \begin{tabular}{llll}
\begin{tabular}{p{1.4cm}llp{1.8cm}}
    \hline
  \textbf{Dataset} & \textbf{System} & \textbf{SARI} $\uparrow$ & \textbf{FKLG} $\downarrow$ \\
  \hline  \hline \multicolumn{4}{c}{Development Set} \\ \hline 
  Wikilarge/   & Cosine            & 43.49                    & 7.69                       \\   TurkCorpus          & BART            & 48.81                    & 6.91                       \\
    & NC-TS           & 49.06 (+0.3)              & 6.23 (-0.7)                 \\
                     & Oracle          & 58.09 (+9.3)              & 6.57 (-0.3)                 \\
  \hline
  Wikismall  & Cosine            & 40.99                    & 8.58                       \\        & BART            & 44.25                    & 7.79                       \\
                     & NC-TS           & 46.86 (+2.6)              & 7.20 (-0.6)                 \\
                     & Oracle          & 54.30 (+10.0)             & 7.72 (-0.1)                 \\
  \hline
  Newsela     & Cosine            & 38.03                    & 5.90                       \\       & BART            & 41.24                    & 4.92                       \\
                     & NC-TS           & 43.63 (+2.4)              & 3.71 (-1.2)                 \\
                     & Oracle          & 52.32 (+11.1)             & 4.43 (-0.5)                 \\ 
  \hline \hline \multicolumn{4}{c}{Test Set} \\ \hline
    Wikilarge/  & Cosine            & 38.38                    & 8.79                       \\ TurkCorpus     & BART            & 39.73                    & 8.10                       \\
          & NC-TS           & \textbf{40.24} (+0.5)     & \textbf{7.25} (-0.9)        \\
               \hline
    Wikismall  & Cosine            & 39.00                    & 11.04                       \\      & BART            & 40.52                    & 10.9                       \\
                     & NC-TS           & \textbf{44.46} (+3.9)     & \textbf{9.63} (-1.3)        \\
                \hline
    Newsela     & Cosine            & 38.12                    & 5.64                       \\     & BART            & 40.19                    & 4.82                       \\
                     & NC-TS           & \textbf{41.67} (+1.5)     & \textbf{3.65} (-1.2)        \\
                \hline

\end{tabular}
\caption{Results on the development dataset (top) and test set (bottom). Best results in test in bold. Numbers in parenthesis represent the gain wrt. BART.}
  \label{tab:results-devel-datas}
\end{table}

We start by showing the results for the development set in the upper
part in Table \ref{tab:results-devel-datas}. The oracle system reveals
that selecting the first candidate from BART is not always the best
strategy, and that there is ample room for improvement if we manage to
correctly choose the best candidate among the top $10$
alternatives. However, a straightforward baseline like Cosine does not
improve BART results. The NC-TS system does improve BART results, but this is
expected, as the hyperparameter search is performed in this
development set. The $\lambda$ values that yielded best
results\footnote{$(\lambda_1,\lambda_2, \lambda_3,\lambda_4)$:
  TurkCorpus $(0.5, 0.0, 0.1, 0.6)$, WikiSmall $(0.5, 0.0, 0.1, 0.6)$,
  Newsela $(0.9, 0.2, 0.7, 0.1)$.} show that $p(y|x)$ is a reliable
indicator, but that the contribution of $p(x|y)$ is almost
negligible. The language model $p(y)$ barely helps in Turkcorpus and
Wikismall, but this is expected as these datasets are based on simple
Wikipedia, the same corpus $p(y)$ is trained on. However, the system
assigns a high score to $p(y)$ in the Newsela dataset, showing that
the additional information provided by simple Wikipedia helps
improving the results.

\begin{table}[t]
  \begin{center}
  \resizebox{1.05\columnwidth}{!}{%
    \begin{tabular}{lll}
      \hline
      \textbf{Systems}                                  & \textbf{SARI}$\uparrow$ & \textbf{FKLG}$\downarrow$ \\
      \hline \multicolumn{3}{c}{Wikilarge / Turkcorpus}\\\hline
      DRESS-LS  \cite{zhang-lapata-2017-sentence}       & 37.27                   & 6.62                      \\
      NSELSTM-S \cite{vu-etal-2018-sentence}            & 36.88                   & ---                       \\
      DMASS \cite{zhao-etal-2018-integrating}           & 40.45                   & 8.04                      \\
      EditNTS \cite{dong-etal-2019-editnts}             & 38.22                   & 7.30                      \\
      ACCESS \cite{martin-etal-2020-controllable}       & 41.87                   & 7.22                      \\

      MUSS \cite{martin2020multilingual} & 42.53                   & 7.60                      \\

      SDISS \cite{Lin_Wan_2021}                             & 38.66          & 7.07  \\
      Edit+synt \cite{cumbicus-pineda-etal-2021-syntax}     & 36.97          & 7.46  \\
      TST \cite{Omelianchuk}                                & 41.46          & 7.87  \\
      Control Prefixes \cite{Clive-2021}                    & 42.32          & 7.74  \\
      TS\_T5 \cite{sheang-saggion-2021-controllable} & \textbf{43.31} & 
                                                \textbf{6.17}                        \\
      T5+control \cite{sanja-saggion-2022}                  & 43.30          & ---   \\
      \hline

      NC-TS                                                & 40.24          & 7.25  \\ \hline
      \multicolumn{3}{c}{WikiSmall}                                                \\\hline
      DRESS-LS \cite{zhang-lapata-2017-sentence}         & 27.24          & 7.55          \\
      NSELSTM-S \cite{vu-etal-2018-sentence}             & 29.75          & ---           \\
      EditNTS \cite{dong-etal-2019-editnts}              & 32.35          & 5.47          \\
      SDISS \cite{Lin_Wan_2021}                          & 34.06          & \textbf{4.58} \\
      TST \cite{Omelianchuk}                             & \textbf{44.67} & 9.29          \\
      \hline

      NC-TS                          & 44.46                   & 9.63                      \\ \hline
      \multicolumn{3}{c}{Newsela}                                             \\\hline
      DRESS-LS \cite{zhang-lapata-2017-sentence}            & 26.63          & 4.21  \\
      NSELSTM-S \cite{vu-etal-2018-sentence}                & 29.58          & ---   \\
      DMASS \cite{zhao-etal-2018-integrating}               & 27.28          & 5.17  \\
      
      EditNTS \cite{dong-etal-2019-editnts}              & 31.41          & 3.40          \\

      MUSS \cite{martin2020multilingual}                 & 41.17          & 2.70          \\
      SDISS \cite{Lin_Wan_2021}                          & 32.30          & \textbf{2.38} \\
      EditSynt \cite{cumbicus-pineda-etal-2021-syntax}  & 38.08          & 4.60          \\
      \hline
     
      NC-TS                                             & \textbf{41.67} & 3.65          \\ \hline

    \end{tabular}%
    }

    \caption{Comparison of our system with state-of-the-art neural systems, values have been obtained from the respective papers.}
    \label{tab:comparison-with-state}
  \end{center}
\end{table}

The main results on the test set are shown in the bottom part in Table
\ref{tab:results-devel-datas}. The NC-TS system obtains the best results on
both metrics on all datasets, which confirms the previous results on
development, and shows that the method is robust across datasets. In Table
\ref{tab:comparison-with-state} we compare NC-TS with state-of-the-art
systems. Our results are competitive with the state-of-the-art systems and
in Newsela we obtain the best results in the SARI metric and a good score in
FKLG.

\section{Analysis}
\label{sec:analysis}

In this section we present the manual evaluation we have carried out to compare BART against the NC-TS system. 

We first conducted a manual A/B test comparing the systems in the WikiSmall dataset, the one where the gain of NC-TS is the highest. We sampled 25 sentences from the output of each system, stratified according to the length of the original complex sentences\footnote{We grouped the sentences into four groups according to their length, and randomly sampled about 11-13 sentences on each group.}. Two linguists with expertise in ATS have evaluated the sample. They were not English native speakers, but had advanced studies (one B.A. degree, the other a Phd.). They were presented the complex sentence and the outputs of both systems, and they had to choose one of them. They were allowed to evaluate both sentences as equal, but were instructed to do so only in very clear cases. When evaluating the sentences, they did not know which systems they belonged to, neither did they know the reference simplification in the gold standard. They followed the following criteria in the evaluation:

\begin{itemize}
\item Try always to always choose one system, and use equality only on extreme cases. 
\item Penalize hallucinations and misinformation, since they are known to cause harm \cite{cumbicus2021linguistic}.
\item All other criteria being equal, select the shortest sentence.
\item Prioritise the meaning preservation and those sentences that keep the nuances.

\end{itemize}

In Table \ref{tab:manualWikiSmall} we present the results of the manual evaluation, distributed in different sentence lengths. In total, the NC-TS system has obtained the best results ($30$ sentences) and the difference is higher on longest sentences (from quartile 2 on), which implies that the re-ranking is more beneficial on those cases.

\begin{table}[t]
%\small
\centering
%\label{tab:quartiles}{
\begin{tabular}{lllll|l}
\hline
  Quartile      & Q1 & Q2 & Q3 & Q4 & Total    \\ \hline
  BART          & 5  & 4  & 3  & 3  & {\bf 15} \\  %\hline
  NC-TS         & 6  & 8  & 9  & 7  & {\bf 30} \\ %\hline
  Equal         & 1  & 1  & 1  & 2  & {\bf 5}  \\ \hline
    %     Total & 11 & 13 & 13 & 12 & 50       \\ \hline
\end{tabular}
%}
\caption{Results of the manual evaluation. The Q columns correspond to the quartiles according to the sentence lengths.} 
\label{tab:manualWikiSmall}
\end{table}
We performed an additional analysis on the WikiLarge and Newsela datasets. We selected the $20$ simplifications on each dataset where the SARI difference between Bart and NC-TS was highest ($10$ examples in BART favor, $10$ examples on NC-TS favor)\footnote{Examples of both systems can be found in appendix \ref{sec:simpl-exampl}}. In the cases that NC-TS is better, we found out that the output of BART is mainly a copy of the original sentence, that is, no simplification has been performed. This happens particularly in the WikiLarge dataset, but also in Newsela, although to a lesser extent. In Newsela, the NC-TS outputs clearly show a wider variety of operations (coreference changes, lexical simplification, person changes) compared to BART. On the other hand, when BART scores better, we found out that BART carries out slight changes with respect to the original and NC-TS tends to copy the original complex sentence. Again, this happens mostly in WikiLarge; in Newsela, the casuistry in the analysed dataset is bigger and no general conclusion can be obtained from the evaluated sample.

\section{Conclusion and Future Work}
\label{sec:conclusion}

In this paper we have presented simple re-ranking method to used the noisy channel in Automatic Text Simplification. We have compared our NC-TS system to a state-of-the-art seq2seq model (BART) and shown that the method obtains better results in terms of SARI and the readability metric FKLG in three well known ATS dataset (WikiSmall, WikiLarge/TurkCorpus and Newsela). Moreover, the NC-TS system obtains competitive results compared to state-of-the-art systems, and yields best results in the Newsela dataset. In this dataset, the system is able to leverage new information coming from the simple Wikipedia, which is not part of the dataset itself. We have conducted manual analysis in WikiSmall, where our NC-TS system has the most profit, and show that its outputs are better qualitatively, particularly on longer sentences. 

In the future we plan to test the NC-TS system in ATS datasets in cross-lingual settings, where the models that compose the noisy channel schema can be trained using different datasets on several languages.

\section{Limitations}
\label{sec:Limitations}

Automatic text simplification is a subjective task: it highly depends on the target audience and experts do not agree on a perfect definition for simplification. Based on that, our manual evaluation can be biased towards the criteria explained in this paper of our experts and their experience. Moreover, all manual evaluations are subjective and difficult to reproduce.

This research has only been conducted for English due to the fact of the large data scarcity in other languages. Besides, Newsela, one of the datasets we have used is a proprietary dataset (data can be obtained in \url{https://newsela.com/data/}). No examples can be shown unless explicitly allowed, therefore hindering the reproducibility of the results. 

% \section*{Acknowledgements}

% \todo{poner proyectos}

\bibliography{eacl2023}
\bibliographystyle{acl_natbib}

\appendix

\section{Simplification examples}
\label{sec:simpl-exampl}
In this Appendix we show four simplification examples. The first two examples come from the WikiSmall dataset and the NC-TS outputs were evaluated as better. The third and forth examples come from WikiLarge and the NC-TS scored a higher SARI while in the fifth and sixth examples BART scored better. We cannot legally show examples from Newsela.

\begin{table*}[t]
%\small
\centering
%\label{tab:quartiles}{
\begin{tabular}{p{0.1\textwidth}p{0.3\textwidth}p{0.3\textwidth}p{0.3\textwidth}}%{lll}
\hline
{\bf Dataset} &{\bf Original}  & {\bf BART} & {\bf NC-TS} \\ \hline
WikiSmall &Being much more resistant to cold and moist weather conditions, brick enabled the construction of permanent buildings in regions where the harsher climate precluded the use of mud bricks.& 	being much more resistant to cold and moist weather conditions, brick enabled the construction of permanent buildings in regions where the harsher climate precluded the use of mud bricks. & brick is much more resistant to cold and moist weather conditions. this made it possible to build permanent buildings in regions where the harsher climate meant that mud bricks were not suitable. \\
 &It is therefore pointless to email our contact addresses asking for permission to reproduce articles or images, even if rules at your company or school or organization mandate that you ask web site operators before copying their content.&   It is therefore pointless to email our contact addresses asking for permission to reproduce articles or images, even if rules at your company or school say that you ask web site operators before copying their content. &  do not email our contact addresses asking for permission to reproduce articles or images, even if rules at your company or school say that you ask web site operators before copying their content.\\ \hline
 
 WikiLarge (NC-TS better SARI) & He served as the prime minister of Egypt between 1945 and 1946 and again from 1946 and 1948. & He served as the prime minister of Egypt between 1945 and 1946 and again from 1946 and 1948.	 & He was the prime minister of Egypt from 1945 to 1946 and again from 1946 to 1948.	\\
 
  & a town may be correctly described as a market town or as having market rights even if it no longer holds a market , provided the right to do so still exists. & a town may be described as a market town or as having market rights even if it no longer holds a market, provided the right to do so still exists. &    a town may be described as a market town even if it does not have a market anymore, as long as the right to hold a market still  exists.	\\ \hline
  
 WikiLarge (BART better SARI) &  Disney received a full-size Oscar statuette and seven miniature ones , presented to him by 10-year-old child actress Shirley Temple. &		Disney received a full-size Oscar statuette and seven miniature ones, presented to him by Shirley Temple, who was then 10 years old. & Disney received a full-size Oscar statuette and seven smaller ones, presented to him by 10-year-old child actress Shirley Temple. \\
 
   &  Eugowra is said to be named after the indigenous Australian word meaning '' the place where the sand washes down the hill '' .  & 	Eugowra is said to be named after the Australian word meaning ''the place where the sand washes down the hill''.  &	Eugowra is said to be named after the indigenous Australian word meaning ''the place where the sand washes down the hill''.  \\
 
\hline
\end{tabular}
%}
\caption{Outputs of BART and the NC-TS systems} % For our analysis, we obtained 72 sentences, 24 sentences for each of the datasets, and 6 sentences for each quartile
\label{tab:examplesWikiSmall}
\end{table*}

\end{document}